\pdfoutput=1

\documentclass[11pt]{article}

\usepackage[final]{acl}

\usepackage{times}
\usepackage{latexsym}

\usepackage[T1]{fontenc}

\usepackage[utf8]{inputenc}

\usepackage{microtype}

\usepackage{inconsolata}

\usepackage{graphicx}

\usepackage{amsmath}
\usepackage{multirow}
\usepackage{booktabs}

\title{RoE-FND: A Case-Based Reasoning Approach with Dual Verification for Fake News Detection via LLMs}

\author{Yuzhou Yang \and Yangming Zhou \\
  Fudan University \\
  College of Computer Science and Artificial Intelligence \\
  \texttt{\{yangyz22, ymzhou21\}@m.fudan.edu.cn} \AND
  Zhiying Zhu \\
  East China University \\
  of Science and Technology \\
  School of Information Science and Engineering \\
  \texttt{zhiyingzhu@ecust.edu.cn}
 \And
  Zhenxing Qian \and Xinpeng Zhang \\
  Fudan University \\
  College of Computer Science \\ and Artificial Intelligence \\
  \texttt{\{zxqian, xinpengzhang\}@fudan.edu.cn}
}

\begin{document}
\maketitle
\begin{abstract}
The proliferation of deceptive content online necessitates robust Fake News Detection (FND) systems. 
While evidence-based approaches leverage external knowledge to verify claims, existing methods face critical limitations: noisy evidence selection, generalization bottlenecks, and unclear decision-making processes. 
Recent efforts to harness Large Language Models (LLMs) for FND introduce new challenges, including hallucinated rationales and conclusion bias. 
To address these issues, we propose \textbf{RoE-FND} (\textbf{\underline{R}}eason \textbf{\underline{o}}n \textbf{\underline{E}}xperiences FND), a framework that reframes evidence-based FND as a logical deduction task by synergizing LLMs with experiential learning. 
RoE-FND encompasses two stages: (1) \textit{self-reflective knowledge building}, where a knowledge base is curated by analyzing past reasoning errors, namely the exploration stage, and (2) \textit{dynamic criterion retrieval}, which synthesizes task-specific reasoning guidelines from historical cases as experiences during deployment.
It further cross-checks rationales against internal experience through a devised dual-channel procedure. 
Key contributions include: a case-based reasoning framework for FND that addresses multiple existing challenges, a training-free approach enabling adaptation to evolving situations, and empirical validation of the framework’s superior generalization and effectiveness over state-of-the-art methods across three datasets. 
\end{abstract}

\section{Introduction}

Contemporary media platforms, such as news feeds and microblogs, are witnessing a growing prevalence of deceptive and manipulative material.
This includes dubious assertions, "alternative facts'', or even entirely fabricated news stories~\cite{shu2017fake, fisher2016pizzagate}. 
The proliferation of such content erodes public trust and exacerbates societal polarization, making automated Fake News Detection (FND) systems a critical line of defense~\cite{shu2020fakenewsnet}.
Early methods relied on shallow textual features like lexical statistics~\cite{castillo2011information} or syntactic patterns~\cite{feng2012syntactic}.
Recent works that based on deep-learning techniques prepare adequate news samples from the real world, including authentic news and those suspicious or fabricated~\cite{shu2020fakenewsnet, popat2017truth, jin2017multimodal}. Researchers can thus develop methods to capture semantic patterns of deception~\cite{zhang2021mining, kaliyar2021fakebert, zhu2022memory}.
However, these approaches lack support from related factual information, limiting their adaptability to evolving manipulation tactics.

\begin{figure}[!t]
  \centering
  \includegraphics[width=\columnwidth]{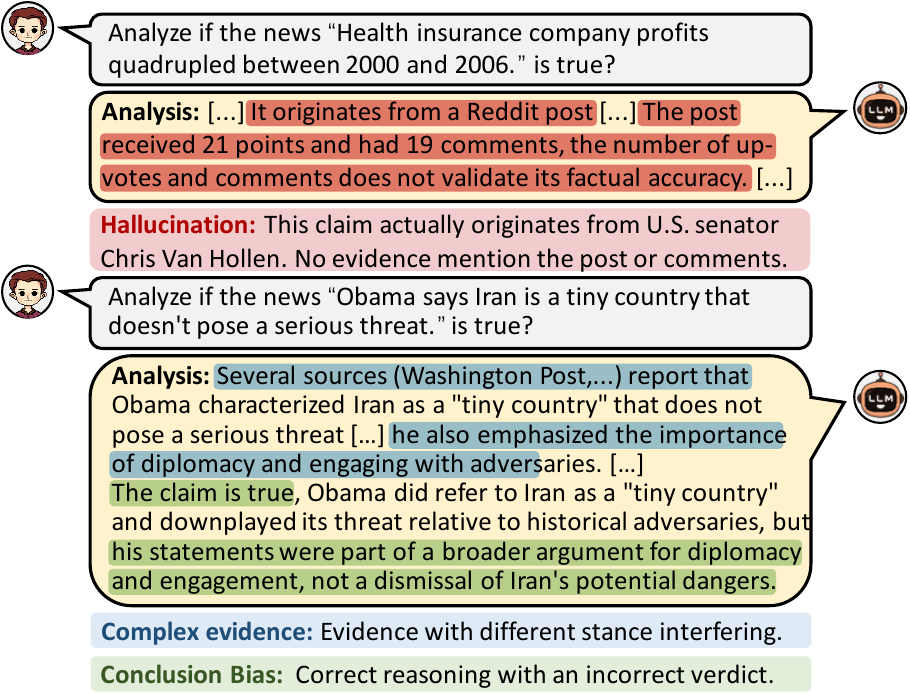}
  \caption{
 Examples of common mistakes made by LLMs when analyzing the news's authenticity. We omit the less important content using [...].}
  \label{fig:intro}
\end{figure}

This challenge has spurred interest in evidence-based FND, where models verify claims against external knowledge sources~\cite{li2016survey, wu2021unified}.
Current approaches mainly employ two strategies: 1) Evidence filtering via lexical similarity~\cite{rashkin2017truth} or neural retrievers~\cite{yang2022coarse}, and 2) Learning joint representations of news-evidence pairs~\cite{ma2019sentence, wu2021unified}.
While these methods achieve domain-specific success, our analysis reveals critical limitations.
Firstly, current evidence selection mechanisms lack explicit reasoning traces, making them vulnerable to noisy or adversarial evidence.
Secondly, models are trained on platform-biased or domain-specific datasets, this brings a generalization bottleneck that causes performance drops when tested on news from different domains or platforms~\cite{zhu2022memory}.
Thirdly, the interpretability gap brought by end-to-end neural architectures obscures the decision logic, hindering practical deployment in high-stakes scenarios.

Large Language Models (LLMs) offer promising potential for addressing these challenges with their emergent textual reasoning capabilities~\cite{wei2022emergent} and zero-shot generalization ability that requires no model parameters tuning~\cite{kojima2022large}.
Recent attempts directly employ LLMs as fact-checkers~\cite{pan2023fact} or claim verifiers~\cite{caramancion2023news}, yet introduce new challenges.
For instance, LLMs may generate plausible but factually incorrect rationales, a phenomenon known as hallucinated reasoning~\cite{huang2023survey}.
Besides, harder than fact-checking tasks that only require determining the evidence's stance~\cite{thorne2018fever}, evidence-based FND via LLMs often struggle with complex evidence synthesis that may contain supporting, refuting, and misguiding materials.
Thirdly, recent research reveals the conclusion bias problem where correct reasoning inexplicably leads to incorrect verdicts~\cite{hu2024bad}. 
We present two examples in Figure~\ref{fig:intro} illustrating these challenges.
These limitations highlight the need for a principled framework that can harness the reasoning power of LLMs while mitigating their vulnerabilities.

In this paper, 
we reframe evidence-based FND as a logical deduction task and present a framework, namely RoE-FND (\underline{\textbf{R}}eason \underline{o}n \underline{\textbf{E}}xperiences FND), that employs LLMs as logical reasoning units and synergize their power with case-based experiential learning.
Unlike prior LLM-based approaches that perform one-off verification~\cite{pan2023fact, caramancion2023news}, RoE-FND comprises two stages and introduces several key innovations: 1) \textit{Self-reflective experience building}: during the exploration stage, the model constructs a knowledge base through self-questioning the wrong answer. 
2) \textit{Dynamic criterion retrieval}: at the deployment stage, task-specific advice is dynamically synthesized from relevant historical cases, assisting in finding better rationales.
3) \textit{Dual-channel verification mechanism}: RoE-FND cross-checks generated rationales against both external evidence and internal experience patterns.

We conduct extensive experiments on three challenging datasets, i.e. CHEF~\cite{hu2022chef}, Snopes~\cite{popat2018declare}, and PolitiFact~\cite{shu2020fakenewsnet} to validate the effectiveness of RoE-FND.
The results demonstrate improvements in various metrics over state-of-the-art methods.
We also propose a fine-tuning strategy for RoE-FND, which brings significant improvements.
Quantitative results and analysis of generated content underscore the advantages of our design, particularly in terms of robustness, interpretability, and generalization.
In summary, our work makes several key contributions:

\begin{itemize}
    \item We propose RoE-FND, formalizing experiential reasoning for FND through a case-based strategy. It can produce accurate explanations for predictions of the news's authenticity.
    \item We devise a training-free strategy that incorporates self-reflective experience curation and dynamic criterion adaptation, offering a novel approach to leveraging LLMs for FND.
    \item Extensive experiments on multiple settings validate the advantages of our framework. 
    Detailed studies offer insights into the mechanisms driving the framework's success.
\end{itemize}

\begin{figure*}[!t]
  \centering
  \includegraphics[width=1.0\textwidth]{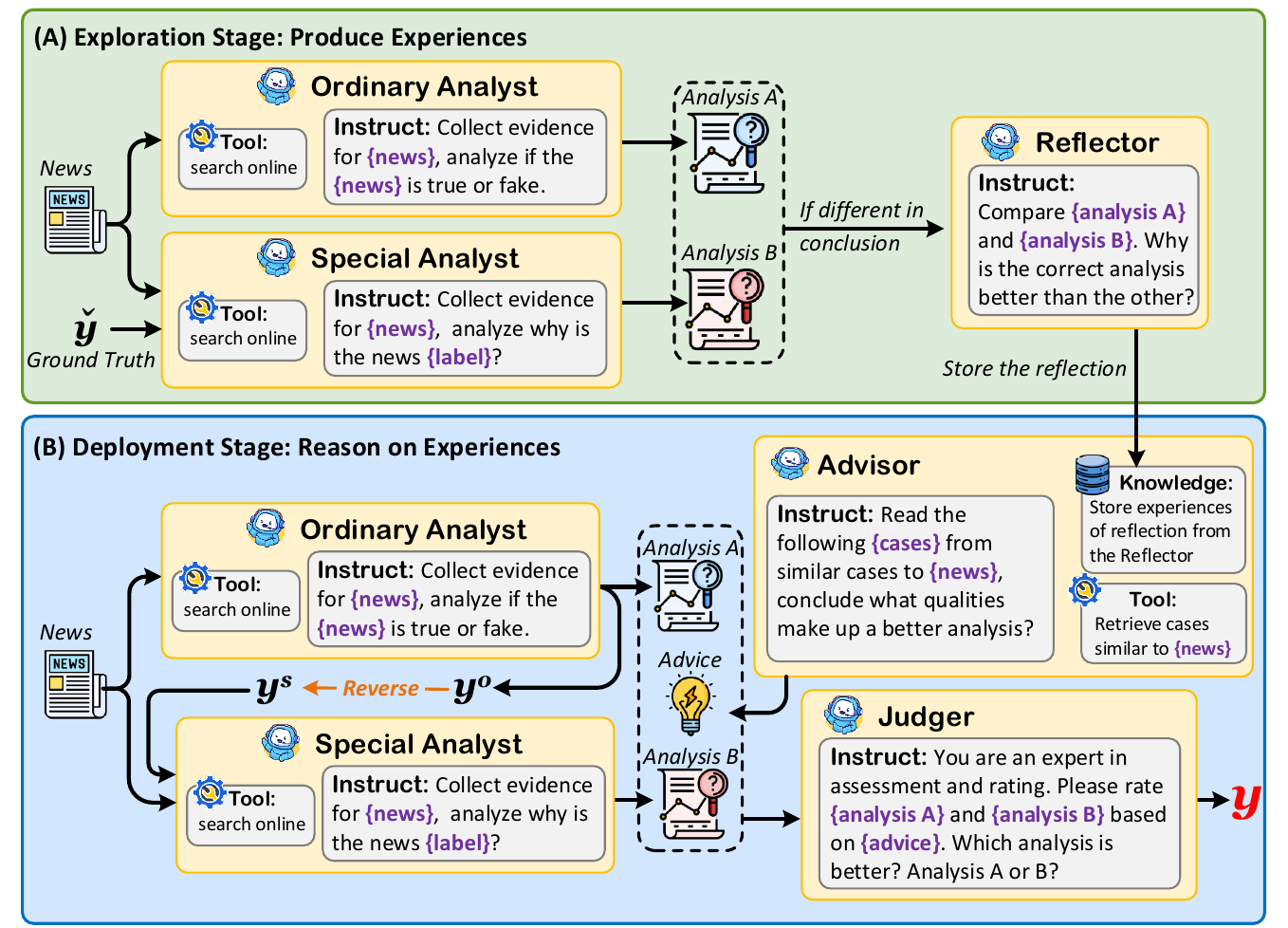}
  \caption{
  Framework design of \textbf{RoE-FND}.
\textbf{Exploration Stage}: construct a knowledge base through self-reflective experience building.
\textbf{Deployment Stage}: dynamically synthesizes advice from historical cases to enhance detection.}
  \label{fig:architecture}
\end{figure*}

\section{Related Works}
\label{sec:related_works}

\subsection{Evidende-based Fake News Detection}
Evidence-based fake news detection conducts knowledge comparison between news and relevant evidence materials~\cite{zhou2020survey}. 
Retrieved materials are usually without further relevance checking. Famous datasets of the field, Snopes~\cite{popat2017truth} and PolitiFact~\cite{rashkin2017truth} comply with this retrieval pipeline using Microsoft Bing API to retrieve evidence.
Many customized detection approaches have been suggested, building upon these datasets. 
DeClarE~\cite{popat2018declare} proposes the earliest evidence-based FND method, it jointly learns representations of news content with evidence materials.
Many works leverage the merits of hierarchical attention for evidence-news interaction, e.g., HAN~\cite{ma2019sentence}, EHIAN~\cite{wu2021evidence}, MAC~\cite{vo2021hierarchical}.
GET proposes a graph neural network to model distant semantic correlation among news and evidence materials~\cite{xu2022evidence}.
MUSER retrieves the key evidence information for news verification in a multi-round retrieval process~\cite{liao2023muser}.
SEE builds a dynamic network to adaptively exploit the coarse evidence materials~\cite{yang2024see}.

\subsection{Large Language Model Inference}
Large language models (LLMs) are trained on extensive corpora and are designed to align with human preferences~\cite{openai2022chatgpt, touvronllama, anthropic2024claude}.
LLMs rely on devised prompts to solve tasks. 
Advanced prompting methods enhance the reasoning abilities of LLMs.
Chain-of-Thought (CoT) achieved considerable improvements by segmenting tasks~\cite{wei2022chain}.
CoT-SC searches for the best solution within multiple traces of CoT~\cite{wang2022self}.
ReAct prompts the LLMs to work in a paradigm of ``observe, thought, then acting'', it achieves better results in complex tasks~\cite{yao2022react}.
ReWOO improves ReAct's solution by asking the LLMs to plan thoroughly before acting~\cite{xu2305rewoo}.
Several studies offer fresh perspectives by equipping LLMs with experiences of similar cases during inference~\cite{yang2023end, sourati2023case, guo2023prompting}.
Therefore, we propose to coordinate advanced inference techniques for our framework. 
Especially, utilizing experiences of similar cases to address the aforementioned challenges of employing LLMs for FND.

\section{Proposed Approach}

In this section, we detail the methodology of the proposed framework RoE-FND, which leverages LLM as reasoning units to solve evidence-based fake news detection tasks.
As shown in Figure~\ref{fig:architecture}, RoE-FND operates in two stages: the exploration stage and the deployment stage. 
Different components are implemented via calling LLM with different task instructions and tool usage.

\subsection{Exploration: Produce Experiences}

In the exploration stage, we structure the workflow of the RoE-FND to produce experiences by exploring samples in the training set.
Directly instructing LLMs to complete the exploration job in one response is unrealistic due to the job's complexity.
Therefore, we design sub-tasks to achieve the goal.

\subsubsection{Step 1: Dual-channel Analyzing} 
In the exploration stage, we assign \texttt{Analyzer} to analyze the news's authenticity.
The \texttt{Analyzer} is equipped with a search tool, which enables it to retrieve factual information by querying the news in search engines during its rationale.
It concludes the news's authenticity by comprehending its analysis, obtaining a predicted label $y\in\left\{true, false\right\}$.

Although the \texttt{Analyzer} can produce step-by-step rationale to compose its analysis, it still risks hallucinated reasoning.
Thereby, we harness the wrong analysis or hallucinated reasoning to produce experiences.
We reveal the label of the news to a special \texttt{Analyzer} ahead, thus it ensures a correct conclusion and possibly has a correct analysis.
By imposing objectives to the LLM, it improves the generation's faithfulness~\cite{dhuliawala2024chain}.

Thus, the news is fed into a dual-channel procedure, with an ordinary and a special \texttt{Analyzer} each. We represent the procedure as:
\begin{align}
\begin{split}
\ddot{y}, \ddot{\mathcal{A}} &= \texttt{OriginalAnalyst}\left(\mathcal{N}\right), \\
\tilde{y}, \tilde{\mathcal A} &= \texttt{SpecialAnalyst}\left(\mathcal{N}, \check{y}\right),
\end{split}
\label{eq:dual_channel}
\end{align}
where $\check{y}\in\{true, false\}$ denotes the ground truth label of the news $\mathcal{N}$, ideally $\tilde{y} = \check{y}$.
The \texttt{Analyzer} produces a predicted label and analysis as output.



\subsubsection{Step 2: Comparative Reflection}
Previous studies have demonstrated the self-reflection abilities of LLMs~\cite{luo2023zero, pan2023automatically}, which can notice the hallucination or logical inconsistency within LLM generations.
However, they normally directly ask LLM to examine the content that needs reflection~\cite{asaiself, jeong2024adaptive}.
Different from them, in RoE-FND, we innovate to boost the reflection by giving the \texttt{Reflector} agent with vanilla analysis $\mathcal{A}$ and crafted analysis $\tilde{\mathcal{A}}$ for comparisons.

Specifically, we locate samples that are erroneously analyzed in $\ddot{\mathcal{A}}$ but correct in $\tilde{\mathcal{A}}$, i.e., satisfy $\ddot{y}\ne\tilde{y}=\check{y}$ in the dual-channel procedure.
It is reasonable to assume that, for these samples, the LLM makes logic errors or hallucinations in $\ddot{\mathcal{A}}$, thereby we delegate \texttt{Reflector} to identify the mistakes within $\ddot{\mathcal{A}}$ while providing it with $\tilde{\mathcal{A}}$ as comparison and reference. We have:
\begin{align}
\begin{split}
\mathcal{R} = \texttt{Reflector}\left(\mathcal{A}, \hat{\mathcal{A}}\right),
\end{split}
\end{align}
where all experiences $\mathcal R$ are stored for future reference in the deployment stage.
\subsection{Deployment: Reason on Experiences}

Our goal during the deployment stage is to enhance LLM's detection fidelity by leveraging reflection experiences in the exploration stage.
The methodology is inspired by the paradigm of utilizing old experiences to understand and solve new problems~\cite{kolodner1992introduction}. 

\subsubsection{Step 1: Variant Dual-channel Analyzing}

In the deployment stage, RoE-FND firstly employs almost the same dual-channel analyzing procedure as Equation~\ref{eq:dual_channel}. 
However, due to the unavailability of the news's ground truth label, in this dual-channel analyzing procedure, we first generate ordinary analysis and prediction $y^o$.
Then, we reverse the ordinary prediction to have an opposite predicted label $y^s$ for the special \texttt{Analyst}. The procedure by step can be formulated as below:
\begin{align}
\begin{split}
y^o,\mathcal{A}^o&=\texttt{OridinaryAnalyst}(\mathcal{N})\\
y^s &= \begin{cases}
true, \emph{\space\space if \space} y^o = false, \\
false, \emph{\space\space if \space} y^o= true.
\end{cases}\\
\tilde y^s,\mathcal{A}^s&=\texttt{SpecialAnalyst}(N,y^s).
\end{split}
\end{align}

\subsubsection{Step 2: Advice Generation}

In this step, we devise an agent \texttt{Advisor} with a knowledge base and a tool to retrieve information from it.
\texttt{Advisor}'s job is to read the experiences from similar cases within the knowledge base, then generalize advice for its colleague to determine the analyses from the dual-channel procedure.

Specifically, given news article $\mathcal{N}$ for detection, \texttt{Advisor} retrieves similar news cases in the knowledge base constructed by storing experiences of the \texttt{Reflector}'s work and retrieves these cases' reflection.
Let the $k$-th entry within the knowledge base be $(\mathcal N_k,\mathcal R_k)$, \texttt{Advisor} utilize a retrieval tool that based on semantical similarity calculation to retrieve $n$ cases.
\texttt{Advisor} then instructed to comprehend these cases and provide advice.
Assume it retrieve $n$ cases from the knowledge base, the generation of advice $\mathcal{C}$ can be presented as:
\begin{align}
\begin{split}
\mathcal{C} &= \texttt{Advisor}\left(\mathcal{N},\left[\mathcal R_1,\dots,\mathcal R_n\right]\right), \\
&\emph{s.t.\space}{\mathcal R_k}\in S=\{(\mathcal N_1,\mathcal R_1),\dots,(\mathcal N_n,\mathcal R_n)\}\\
&\emph{\space\space\space\space\space}S=\underset{{S\subseteq\mathcal{D},\lvert S\rvert=n}}{\arg{\max}}\sum_{(\mathcal N_x,\mathcal R_x)\in S}sim(\mathcal N_x,\mathcal N),\\
\end{split}
\end{align}

\noindent where $\mathcal D$ denotes the knowledge base, $sim(\cdot)$ denotes the similarity score of two news articles calculated by the retrieval tool.

\subsubsection{Step 3: Determine the Better Analysis}

\texttt{Advisor} generates advice by comprehending experiences of similar cases that are harvested by cross-comparing erroneous and correct analyses with reflection, the advice thereby can guide the LLM to identify better analysis from the dual-channel procedure.

In this step, we assign \texttt{Judger}, who is expertise in rating and assessment, to select the correct analysis while retaining advice from \texttt{Advisor}.
\texttt{Judger} critically compares two analyses with opposing conclusions and finally determines which analysis is better.
The chosen analysis's conclusion is regarded as the prediction result.
\texttt{Judger}'s procedure to output the final prediction is represented as:
\begin{align}
\begin{split}
\mathcal{J} &= \texttt{Judger}\left(\mathcal{A}^o, \mathcal{A}^s,\mathcal{C}\right),\\
y&= \begin{cases}
{y}^o, \emph{\space\space if \space} \mathcal{J} \emph{\space indicates\space} \mathcal{A}^o\geq\mathcal{A}^s, \\
\tilde y^s, \emph{\space\space otherwise},
\end{cases}
\end{split}
\end{align}
where we utilize the symbol $\ge$ to represent that  \texttt{Judger} determines the original analysis is better than the specially crafted analysis.


\section{Experiment}
\subsection{Experimental Setups}
\noindent\textbf{Dataset Preparation.} 
We employ three famous datasets. CHEF~\cite{hu2022chef} is a Chinese dataset collected from the real world with multiple domains of news. 
PolitiFact~\cite{rashkin2017truth}, and Snopes~\cite{popat2018declare} for our experiments are collected from fact-checking websites.
We follow previous works' settings to split the dataset, detailed statistics are presented in Table~\ref{tab:datasets}.

\begin{table}[!t]
\centering
  \begin{tabular}{lrrr}
  \toprule
    \multicolumn{1}{l}{\textbf{Dataset}} & \textbf{\# Train} & \textbf{\# Val} & \textbf{\# Test} \\ \midrule
    \multirow{1}{*}{{CHEF}} & 5,754 & 666 & 666 \\
    \multirow{1}{*}{PolitiFact} & 1,919 & 631  & 650 \\
    \multirow{1}{*}{Snopes} & 2,604 & 869 & 868 \\
    \bottomrule
  \end{tabular}
  \caption{Statistics of the divided FND datasets, where the symbol ``\#" denotes "the number of".}
  \label{tab:datasets}
\end{table}

\noindent\textbf{Tools modules.}
RoE-FND involves a search tool for collecting evidence and a tool to retrieve cases within the knowledge base.
For fair comparisons with baselines, we replace the searched results with evidence within datasets.
The retrieval tool utilizes all-MiniLM-L6-v2~\footnote{\url{https://huggingface.co/sentence-transformers/all-MiniLM-L6-v2}} embedding. We retrieve one case from the knowledge base.


\noindent\textbf{Implementation details.}
We utilize the OpenAI platform~\footnote{\url{https://platform.openai.com/}} and DeepSeek platform~\footnote{\url{https://platform.deepseek.com/}} to run RoE-FND with their models.
The LLaMa-series and Qwen-series are run locally using 8 NVIDIA RTX4090 GPUs with Ollama~\footnote{\url{https://ollama.com/}}.


\begin{table*}[!t]
\small
\centering
\begin{tabular}{lllll|llll|llll}
\toprule
\multirow{2}{*}{\textbf{Method}} & \multicolumn{4}{c}{\textbf{CHEF}} & \multicolumn{4}{c}{\textbf{Snopes}} & \multicolumn{4}{c}{\textbf{PolitiFact}}\\ 
\cmidrule{2-13}
 & ACC & F1 & PR & \multicolumn{1}{l}{RC} & ACC & F1 & PR & \multicolumn{1}{l}{RC} & ACC & F1 & PR & RC \\
\midrule
\multicolumn{5}{l}{\textbf{Trained evidence-based FND methods}} \\
\cmidrule{1-4}
DeClarE & 0.589 & 0.581 & 0.583 & 0.568 & 0.786 & 0.725 & 0.610 & 0.852 & 0.652 & 0.653 & 0.667 & 0.637 \\
EHIAN & 0.571 & 0.600 & 0.583 & 0.516 & 0.828 & 0.784 & 0.617 & 0.882 & 0.679 & 0.676 & 0.686 & 0.675 \\
GET & 0.588 & 0.602 & 0.585 & 0.582 & 0.814 & 0.771 & 0.721 & 0.854 & 0.694 & 0.691 & 0.687 & 0.708 \\
ReRead & 0.789 & 0.776 & 0.826 & 0.745 & 0.816 & 0.714 & 0.652 & 0.789 & 0.693 & 0.681 & 0.711 & 0.718 \\
MUSER & 0.608 & 0.612 & 0.603 & 0.631 & 0.841 & 0.745 & 0.699 & 0.798 & 0.729 & {{0.732}} & 0.735 & 0.728 \\
SEE & 0.763 & 0.776 & 0.751 & 0.802 & 0.824 & 0.786 & \textbf{\underline{0.773}} & 0.845 & 0.706 & 0.705 & 0.688 & 0.724 \\
\midrule
\multicolumn{3}{l}{\textbf{Training-free LLM methods}} \\
\cmidrule{1-3}
4o-mini & 0.740 & 0.773 & 0.687 & 0.883 & 0.745 & 0.452 & 0.604 & 0.361 & 0.575 & 0.395 & 0.698 & 0.275 \\
DeepSeekv3 & 0.780 & 0.784 & 0.790 & 0.778 & 0.736 & 0.253 & 0.712 & 0.154 & 0.470 & 0.631 & 0.485 & 0.901 \\
ProgramFC & 0.694 & 0.708 & 0.723 & 0.697 & 0.741 & 0.619 & 0.542 & 0.723 & 0.678 & 0.684 & 0.725 & 0.741 \\
\midrule
\multicolumn{3}{l}{\textbf{RoE-FND with different LLMs}} \\
\cmidrule{1-3}
GPT-3.5 & 0.643 & 0.676 & 0.618 & 0.745 & 0.711 & 0.585 & 0.503 & 0.698 & 0.665 & 0.603 & 0.758 & 0.500 \\
LLaMa3 & 0.722 & 0.688 & 0.806 & 0.612 & 0.750 & 0.675 & 0.554 & 0.865 & 0.726 & 0.720 & 0.741 & 0.701 \\
Qwen2.5 & 0.828 & 0.827 & 0.846 & 0.784 & 0.824 & 0.747 & 0.644 & 0.888 & \textbf{\underline{0.733}} & 0.509 & \textbf{\underline{0.875}} & 0.359\\
DeepSeekv3 & 0.890 & 0.892 & \textbf{\underline{0.876}} & 0.908 & \textbf{\underline{0.863}} & \textbf{\underline{0.791}} & 0.708 & \textbf{\underline{0.896}} & 0.711 & 0.658 & 0.767 & 0.576 \\
4o-mini & \textbf{\underline{0.891}} & \textbf{\underline{0.893}} & \textbf{\underline{0.876}} & \textbf{\underline{0.911}} & 0.860 & 0.774 & 0.732 & 0.822 & 0.712 & \textbf{\underline{0.735}} & 0.662 & \textbf{\underline{0.827}}  \\
\bottomrule
\end{tabular}
\caption{Performance comparison between our method with baseline methods. We report accuracy (ACC), F1-Macro (F1), precision (PR), and recall (RC)}.
\label{tab:comparison}
\end{table*}

\subsection{Performance Comparisons}


\subsubsection{Comparisons of Different LLM Options}

\begin{figure}[!t]
  \includegraphics[width=\columnwidth]{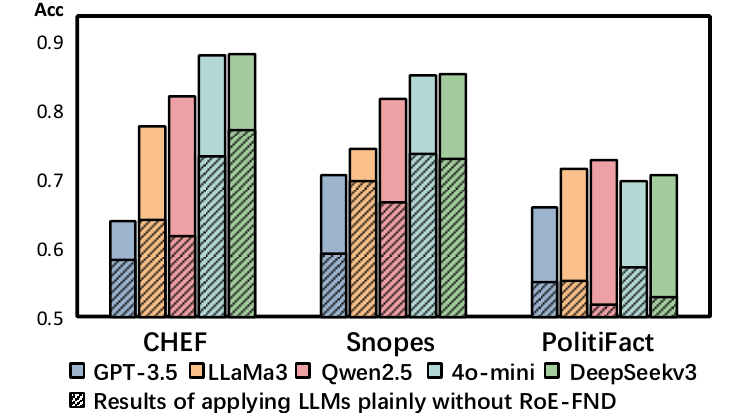}
  \caption{Performance comparisons of RoE-FND utilizing different LLMs on three datasets.}
  \label{fig:model_type}
\end{figure}
\begin{figure}[!t]
  \includegraphics[width=\columnwidth]{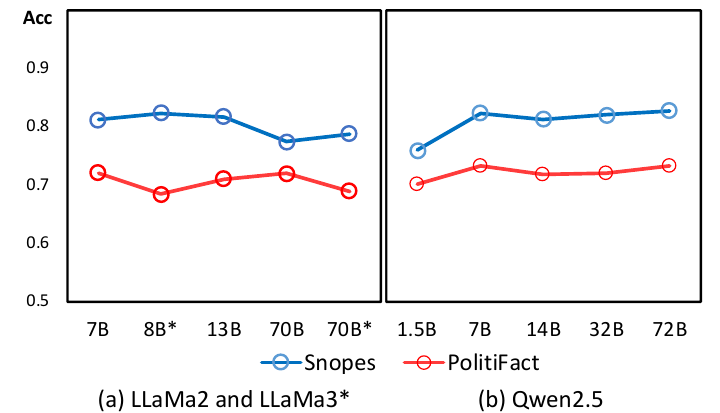}
  \caption{Performance comparison of model size's impact on RoE-FND by two famous models. (a)~Testing of LLaMa2 and LLaMa3 (marked by * symbol). (b)~Testing of Qwen2.5 models.}
  \label{fig:model_size}
\end{figure}

In this section, we evaluate the performance of RoE-FND across various LLM configurations. 
Figure~\ref{fig:model_type} illustrates RoE-FND's detection accuracy using different LLMs. The shaded portion of each bar represents the baseline performance achieved by directly prompting the LLMs without RoE-FND, while the unshaded portion highlights the improvements attributable to RoE-FND. 
Although the extent of improvement varies across datasets, RoE-FND consistently enhances detection accuracy across all LLMs and datasets.
Figure~\ref{fig:model_size} (a) and (b) further explore the impact of model size (under 72B parameters) within two prominent model families. 
In the Qwen2.5 series, the 1.5B model underperforms significantly, likely due to limitations in instruction-following capabilities. 
The 72B model achieves the best performance, despite with only a slight improvement over the 7B model.
The lines in both figures indicate that increasing model size below 72B does not consistently yield performance gains.
Overall, both models exhibit stable detection accuracy for sizes larger than 7B. 

\subsubsection{Comparison with Baseline Approaches}

We compare RoE-FND with multiple methods, including 
DeClarE~\cite{popat2018declare}, EHIAN~\cite{wu2021evidence}, GET~\cite{xu2022evidence}, ReRead~\cite{hu2023read}, MUSER~\cite{liao2023muser}, ProgramFC~\cite{pan2023fact}, and SEE~\cite{yang2024see}.
Details of them are presented in Section~\ref{sec:related_works}.

\begin{table}[!t]
\small
\centering
\renewcommand{\arraystretch}{1.2}
\setlength{\tabcolsep}{1.6mm}
{
  \begin{tabular}{l|l|cc|cc}
  \hline
  \multirow{2}{*}{\centering\textbf{\shortstack{Training \\ Dataset}}} & \multirow{2}{*}{\centering\textbf{Method}} & \multicolumn{2}{c|}{\textbf{Snopes}} & \multicolumn{2}{c}{\textbf{PolitiFact}}
  \\
  \cline{3-6}
  &   & {ACC}  & {$\downarrow$}(\%) & {ACC}  & {$\downarrow$}(\%) \\ 
     \hline

\multirow{4}{*}{Snopes} & GET & 0.814 & - & 0.603 & 42.2 \\ 
& SEE & 0.824 & - & 0.584 & 59.2 \\ 
& MUSER & 0.841 & - & 0.667 & 27.1 \\ 
\cline{2-6}
& \textbf{RoE-FND} & \textbf{0.860} & - & 0.702 & \textbf{4.7} \\
\hline
\multirow{4}{*}{PolitiFact} & GET & 0.692 & 38.8 & 0.694 & - \\ 
& SEE & 0.665 & 49.1 & 0.706 & - \\ 
& MUSER & 0.733 & 31.6 & \textbf{0.729} & - \\ 
\cline{2-6}
& \textbf{RoE-FND} & {0.843} & \textbf{4.7} & 0.712 & - \\
\hline
  \end{tabular}
  }
\caption{Results of cross-datasets testing. {$\downarrow$}(\%) indicates the relative decrease compared to training and testing on the same dataset in percentage.}
\label{tab:crossdatasets}
\end{table}

In Table~\ref{tab:comparison}, we report accuracy (ACC), F1-macro (F1), precision (PR), and recall (RC) following previous methods.
The supervised-trained FND methods exhibit varying levels of performance across the datasets. 
RoE-FND outperforms both supervised-trained evidence-based FND methods and standalone LLMs when integrated with powerful LLMs like DeepSeekv3 and GPT-4o-mini.
This highlights the potential of combining RoE-FND with advanced LLMs to achieve state-of-the-art performance in FND tasks.
RoE-FND achieves the best results on nearly all metrics and datasets, with significant performance improvement on CHEF.
On Snopes and PolitiFact, RoE-FND addresses the detection bias of directly prompting LLMs, while achieving the best detection accuracy.
In conclusion, our experiments demonstrate that RoE-FND can effectively leverage the capabilities of LLMs to improve FND detection across multiple datasets.


\subsubsection{Cross-datasets Testing Performance}

In applications, a trained FND framework is likely to encounter news samples that differ significantly from those in the training dataset. 
Therefore, the generalization capability of a method is crucial for its practical effectiveness. 
We conducted cross-dataset testing to evaluate the generalization ability of RoE-FND. The results, presented in Table~\ref{tab:crossdatasets}, demonstrate that RoE-FND exhibits superior generalization performance compared to baseline methods. This highlights its robustness and adaptability when facing unfamiliar data distributions.

\begin{table}[!t]
\small
\centering
\renewcommand{\arraystretch}{1.2}
\setlength{\tabcolsep}{0.6mm}
{
  \begin{tabular}{l|cc|cc|cc}
  \hline
  \multirow{2}{*}{\centering\textbf{Ablation Setting}} & \multicolumn{2}{c|}{CHEF} & \multicolumn{2}{c|}{Snopes} & \multicolumn{2}{c}{PolitiFact}
  \\
  \cline{2-7}
     & {ACC}  & {$\downarrow$(\%)} & {ACC}  & {$\downarrow$}(\%) & {ACC} & {$\downarrow$}(\%)\\ 
     \hline
\textit{w/o} \texttt{Reflector} & 0.610 & 26.3 & 0.633 & 23.2 & 0.549 & 23.8 \\ 
\textit{w/o} \texttt{Summarizer} & 0.748 & 9.6 & 0.721 & 12.5 & 0.555 & 23.0 \\ 
\textit{w/o} Dual-channel & 0.725 & 12.4 & 0.736 & 10.7 & 0.645 & 10.5 \\ 
\textit{w/o} Case Retrieval  & 0.694 & 16.2 & 0.793 & 3.8 & 0.673 & 6.6 \\ 
\hline
\textbf{Baseline setting} & \textbf{0.828} & \textbf{-} & \textbf{0.824} & \textbf{-} & \textbf{0.721} & \textbf{-} \\ 
    \hline
  \end{tabular}
  }
\caption{Ablation studies of the proposed RoE-FND. {$\downarrow$}(\%) indicates the relative decrease compared to the baseline setting in percentage.}
\label{tab:ablation}
\end{table}
%
\begin{figure*}[!t]
  \includegraphics[width=\linewidth]{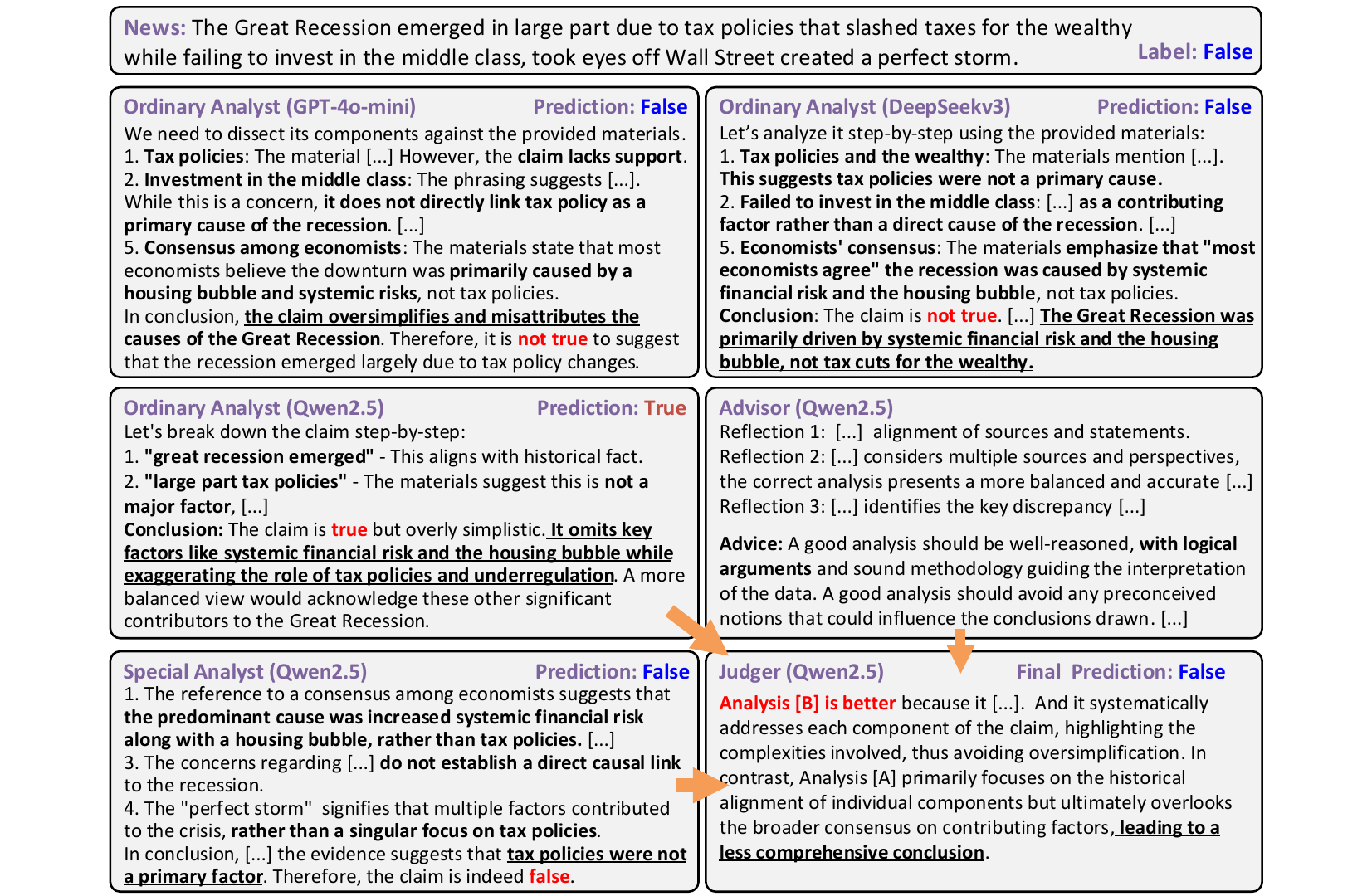}
  \caption{A challenging case from PolitiFact and generation from multiple LLMs. We omit less important content by [..] and highlight key points by colors and boldface.}
  \label{fig:case}
\end{figure*}

\subsection{Framework Design Exploration}

\subsubsection{Ablation Study}
In Table~\ref{tab:ablation}, we investigate the impact of various ablations. 
For \textit{w/o \texttt{Reflector}}, we store analyses without reflection, which are then presented to the \texttt{Judger} as high-quality exemplars, simulating a few-shot learning approach. 
This ablation results in a significant performance drop. 
For \textit{w/o \texttt{Advisor}}, we feed the reflections to the \texttt{Judger} directly. The performance drop underscores the importance of \texttt{Advisor} in maintaining efficiency.
For \textit{w/o Dual-channel}, we replace the dual-channel procedure with a single ordinary \texttt{Analyst}, and without comparative reading during reflection. 
It diminishes the accuracy, demonstrating the value of the dual-channel approach.
For \textit{w/o Case Specified Experience}, we retrieve random cases from the knowledge base. The result emphasizes the necessity of targeted case retrieval for optimal results.

\subsubsection{Enhancement via Fine-tuning}

We devise a solution to enhance the performance of RoE-FND further.
The \texttt{Analyst} is the most critical component of RoE-FND, thereby we propose a fine-tuning strategy that leverages larger LLMs to improve the capabilities of smaller LLMs.
Specifically, we employ a larger LLM, e.g. DeepSeekv3 (671B), to generate high-quality analyses on the training dataset by performing the role of the special \texttt{Analyst}.
These analyses are then used to fine-tune a smaller model, Qwen2.5 (7B), using LoRA (Low-Rank Adaptation). The fine-tuned model serves as the \texttt{Analyst} in RoE-FND, significantly improving its reasoning and analytical capabilities.

We compare the results of enhanced RoE-FND with two state-of-the-art reasoning LLMs: OpenAI o1~\cite{jaech2024o1} and DeepSeek-R1~\cite{guo2025deepseekr1}. The results are presented in Table~\ref{tab:reasoning_model}.
While both reasoning models demonstrate significant improvements over baseline methods, the ordinary RoE-FND still performs competitively, achieving results very close to them. 
Notably, while they take around one minute to process a sample, RoE-FND only takes 22.8s on average.
The RoE-FND with fine-tuning enhancement outperforms both reasoning models on three datasets in detection accuracy. 
The result indicates that enhanced RoE-FND achieves more effective reasoning in FND tasks, leveraging its unique architecture to deliver fast and accurate results.

\begin{table}[!t]
\small
\centering
\renewcommand{\arraystretch}{1.2}
\setlength{\tabcolsep}{1.0mm}
{
  \begin{tabular}{l|cc|cc|cc}
  \hline
  \multirow{2}{*}{\centering\textbf{Method}} & \multicolumn{2}{c|}{\textbf{CHEF}} & \multicolumn{2}{c|}{\textbf{Snopes}} & \multicolumn{2}{c}{\textbf{PolitiFact}}
  \\
  \cline{2-7}
     & {ACC}  & {F1} & {ACC}  & {F1} & {ACC}  & {F1} \\ 
     \hline
OpenAI o1 & 0.892 & 0.888 & 0.852 &\textbf{0.800} & {0.833} & \textbf{0.833} \\
{DeepSeekR1} & 0.882 & 0.880 & 0.820 & 0.791 & 0.800 & 0.792 \\ 
\hline
\textbf{RoE-FND} & 0.891 & {0.893} & {0.863} & 0.791 & 0.702 & 0.735 \\ 
{\space\space+\space\textit{fine-tuned}} & \textbf{0.904} & \textbf{0.908} & \textbf{0.876} & 0.788 & \textbf{0.891} & 0.765 \\
\hline
  \end{tabular}
  }
\caption{Results of advanced reasoning LLMs and RoE-FND enhanced by fine-tuning.}
\label{tab:reasoning_model}
\end{table}

\begin{figure}[!t]
  \includegraphics[width=\columnwidth]{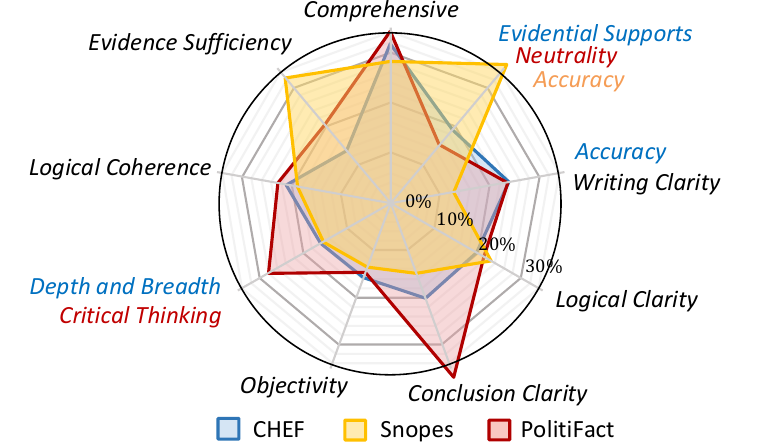}
  \caption{Statistics of the most common keywords from advice by the \texttt{Advisor}. (Black words are shared, colored words are dataset-specified.)}
  \label{fig:advice}
\end{figure}

\subsection{Analysis of Generated Content}


\subsubsection{Case Analysis}

We present a challenging case with corresponding outputs in Figure~\ref{fig:case} to illustrate these findings.
In this example, GPT-4 and DeepSeek-v3 demonstrate a strong ability to dissect the news and analyze each component with supporting evidence.
In contrast, Qwen2.5 follows a similar analytical process but struggles to draw meaningful conclusions due to its overemphasis on descriptive alignment.
The special \texttt{Analyst}, however, achieves more accurate analysis by imposing the conclusion ahead.
\texttt{Advisor} also emphasizes the importance of logical arguments from similar cases, enabling the \texttt{Judger} to recognize the distinctive traits of the analysis and make a correct final prediction.
Through the devised procedure of the RoE-FND framework, a smaller LLM like Qwen2.5 can correctly handle challenging detections as larger LLMs.

\subsubsection{Gains from the Exploration Stage}
\texttt{Advisor} generates suggestions based on insights drawn from similar historical cases
We present the most frequently mentioned keywords from these suggestions in Figure~\ref{fig:advice}.
While there is some overlap in the most common advice keywords across the three datasets, the distribution and emphasis of these keywords vary, likely due to the distinct characteristics of each dataset. 
For example, PolitiFact primarily focuses on news related to policies and political parties, which demands analyses that are neutral, critically constructed, and free from ambiguous conclusions.
In contrast, societal news in Snopes requires rigorous evidential support to verify news.
Despite these differences, stringent standards such as ``comprehensiveness'' remain a consistent requirement.
These findings demonstrate that, even without a formal training process, RoE-FND effectively builds specialized knowledge through its exploration stage.

\section{Conclusion and Future Works}
We introduce RoE-FND, a novel framework that synergies LLMs with case-based experiential learning.
By reframing evidence-based FND as a logical deduction task, RoE-FND leverages self-reflective error analysis to construct a knowledge base, synthesizes advice from historical cases, and ensures the reliability of conclusions through dual-channel verification. 
Experimental results on CHEF, Snopes, and PolitiFact highlight RoE-FND's superior performance in both effectiveness and interpretability compared to existing methods.

Several future directions can further enhance the framework. 
First, incorporating real-time case updates will improve its adaptability in practical applications. 
Second, the framework can be expanded to multi-modal content, e.g. images, to address a broader range of fake news scenarios.
Third, potential biases in LLM-generated rationales can be mitigated through training or human-AI collaboration.
In summary, our work demonstrates the value of integrating experiential learning with logical reasoning, providing a unique and effective approach to tackling the challenges of fake news detection.


\section{Limitations}

While RoE-FND demonstrates promising results in evidence-based FND, several limitations warrant consideration for future improvements.

\noindent\textbf{Reliable evidence availability.}
In RoE-FND, we assign the task of filtering evidence to \texttt{Analyzer} in the deployment stage by inserting an extra step into its action chain before analyzing.
Thereby addressing the problem of noise or biased evidence.
However, the framework may be vulnerable to massive fabricated adversarial evidence materials, which will misguide the LLMs' rationales.

\noindent\textbf{LLM reasoning fidelity.}
The framework assumes that LLMs can generate logically consistent rationales when guided by retrieved criteria. 
However, LLMs are known to suffer from hallucination and contextual bias, which may introduce errors even with dual-channel checks. For example, in our experiments, performance variations were observed across different LLM families (e.g., GPT-4 vs. LLaMa2), suggesting that results are sensitive to the base model’s reasoning capabilities.

\noindent\textbf{Scalability problem.}
The employment of LLMs brings difficulties in scalability.
RoE-FND relies on multiple LLM generations to process each sample, resulting in higher computational latency compared to traditional trained methods. 
Although the framework can be deployed by pipeline, it might be less suitable for applications that require swift decision-making.
Besides, running RoE-FND locally without leveraging LLM APIs may bring a heavy equipment burden.

\noindent\textbf{Potential risks of RoE-FND.}
First, RoE-FND may struggle to detect AI-generated news, as the subtle manipulation traces in such content (e.g., stylistic inconsistencies) are often better identified by trained models with feature extraction. 
Logical reasoning alone may fail to capture these nuances, limiting the framework’s effectiveness against increasingly sophisticated AI-driven misinformation.
Second, although we assume the logical deduction is homogeneous for either language, the cultural generalization in LLMs may be different.
For instance, culturally specific idioms may be neglected during reasoning and cause detection failures.
Moreover, RoE-FND may inherit the bias or ethical shortness of LLMs that caused by training.

Addressing the above limitations is critical for enhancing the robustness, scalability, and practical applicability of the framework in diverse and dynamic misinformation scenarios.

\bibliography{custom}

\appendix

\section{Implementation Details}
\label{sec:appendix}

\begin{figure}[h]
  \includegraphics[width=\columnwidth]{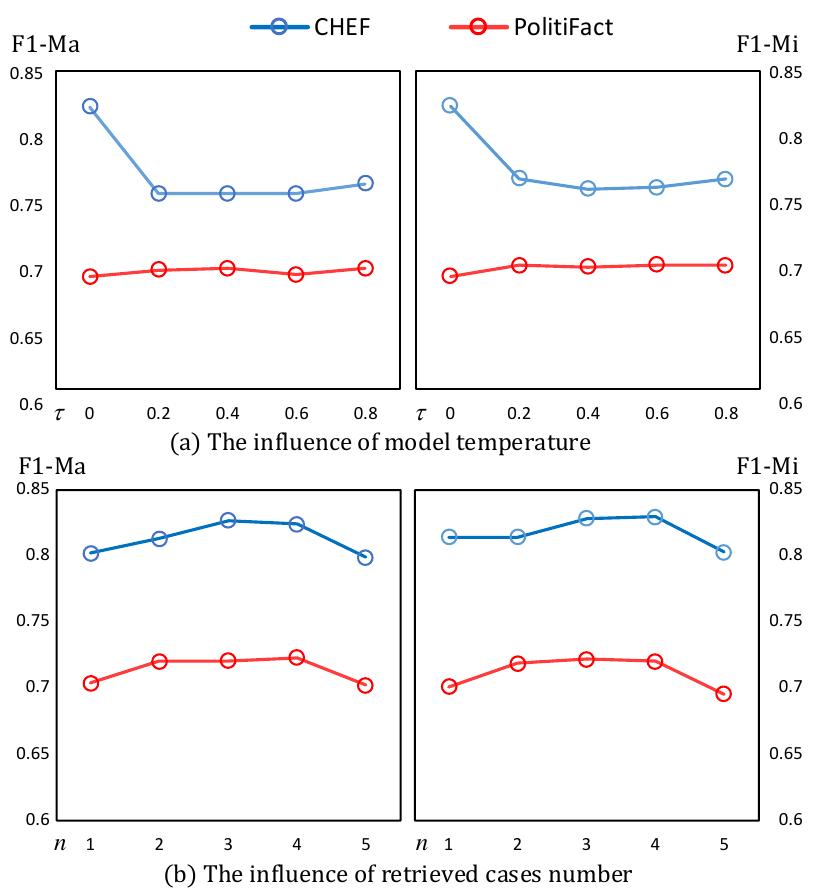}
  \caption{Results of different LLM temperatures and retrieved cases number.}
  \label{fig:hyper}
\end{figure}

\begin{figure*}[!t]
  \centering
  \includegraphics[width=1.0\textwidth]{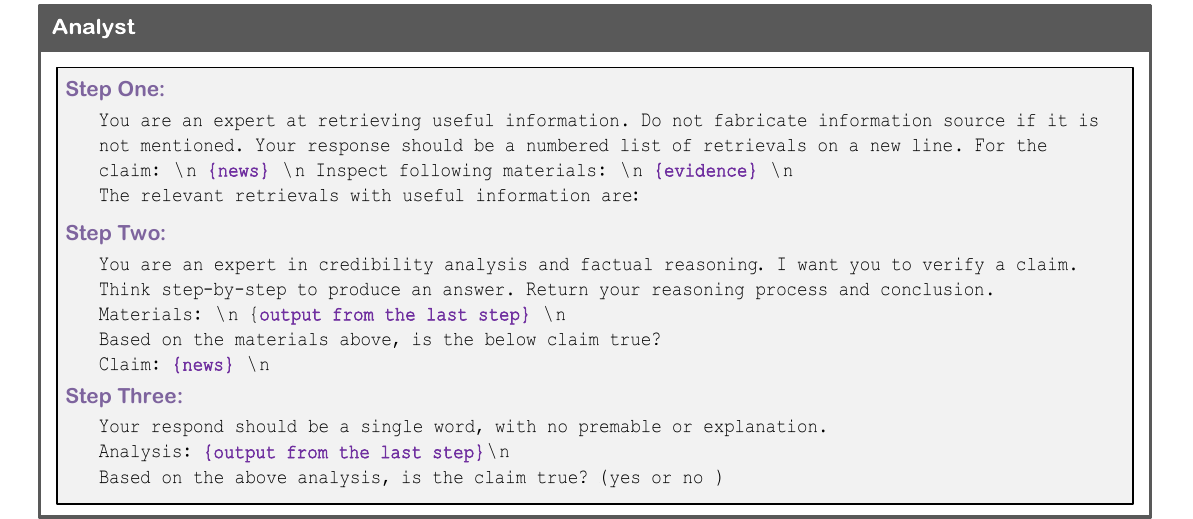}
  \includegraphics[width=1.0\textwidth]{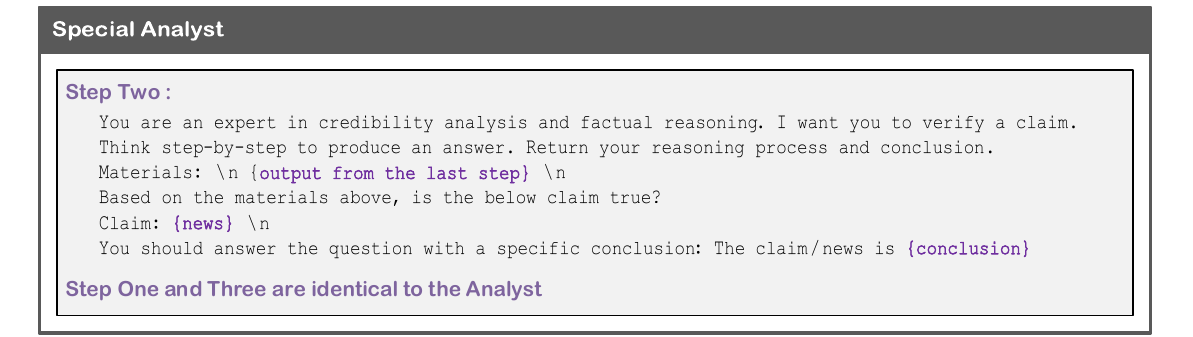}
  \caption{Prompts of ordinary and special \texttt{Analyst}.}
  \label{fig:prompt_analyst}
\end{figure*}

\noindent\textbf{Parameter settings.}
The impact of LLM temperature value and hyperparameter of RoE-FND is depicted in Figure~\ref{fig:hyper}.
For all LLMs employed, we apply the default temperature and top\_p value, the length of generation is set under 2,048 tokens.
Information on model sizes is listed in Table~\ref{tab:licenses}. The Ollama platform uses Q4 quantization.
We utilize DeepSeek and OpenAI models via the officially provided API calls.

\begin{table*}[!t]
\small
\centering
  \begin{tabular}{lrrl}
  \toprule
    \multicolumn{1}{l}{\textbf{Model}} & \textbf{Parameters} & \textbf{Platform} & \multicolumn{1}{c}{\textbf{License}} \\ \midrule
    LLaMa2 & 6.74B, 13B, 69B & Ollama & LLAMA2 COMMUNITY LICENSE \\
    LLaMa3 & 8.03B, 70.6B & Ollama & META LLAMA3 COMMUNITY LICENSE \\
    Qwen2.5 & 1.54B, 7.62B, 14.8B, 32.8B & Ollama & Apache License Version 2. \\
    Qwen2.5 & 72.7B & Ollama & Qwen RESEARCH LICENSE \\
    DeepSeekv3 & 671B & DeepSeek & MIT LICENSE \\
    DeepSeekR1 & 671B & DeepSeek & MIT LICENSE \\
    gpt-3.5-turbo & - & OpenAI & \url{https://openai.com/policies/terms-of-use/} \\
    gpt-4o & - & OpenAI & \url{https://openai.com/policies/terms-of-use/} \\
    gpt-4o-mini & - & OpenAI & \url{https://openai.com/policies/terms-of-use/} \\
    o1-mini & - & OpenAI & \url{https://openai.com/policies/terms-of-use/} \\
    \bottomrule
  \end{tabular}
  \caption{Information of the utilized models.}
  \label{tab:licenses}
\end{table*}

\begin{figure}[!t]
  \centering
  \includegraphics[width=1.0\columnwidth]{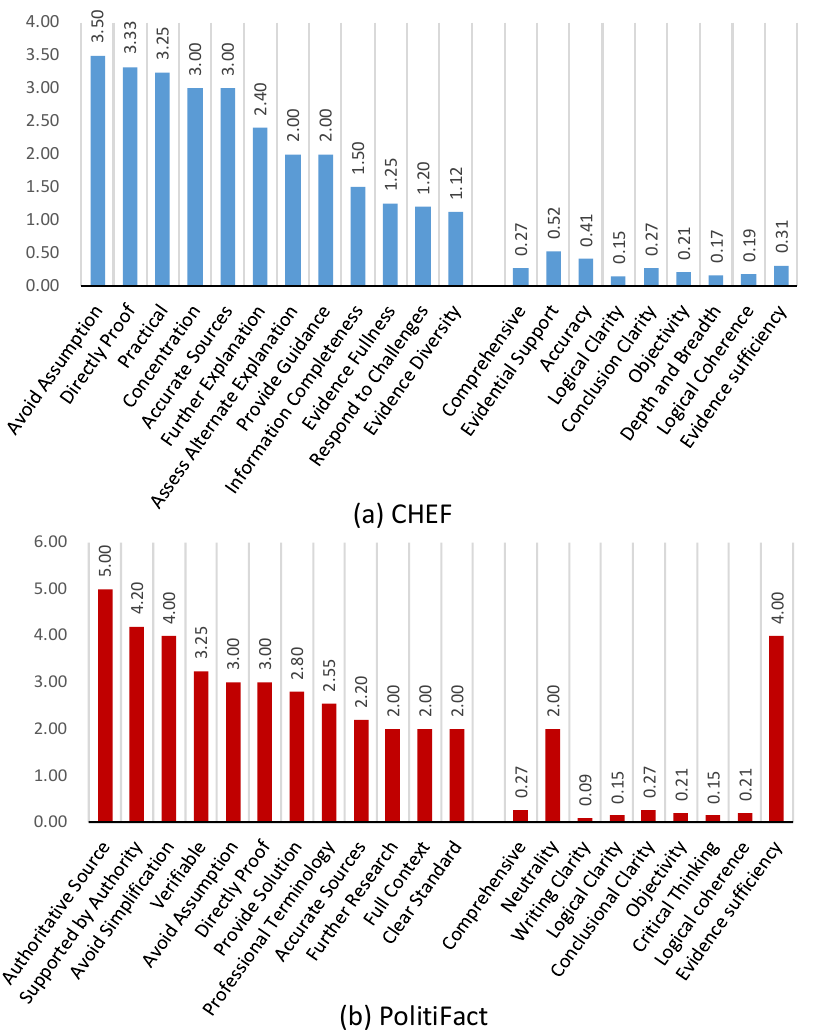}
  \caption{statistics of the most advantageous criteria keywords for the \texttt{Judger} to select the correct analysis. 
  }
  \label{fig:judger_generation}
\end{figure}

\begin{figure*}[!t]
  \centering
  \includegraphics[width=1.0\textwidth]{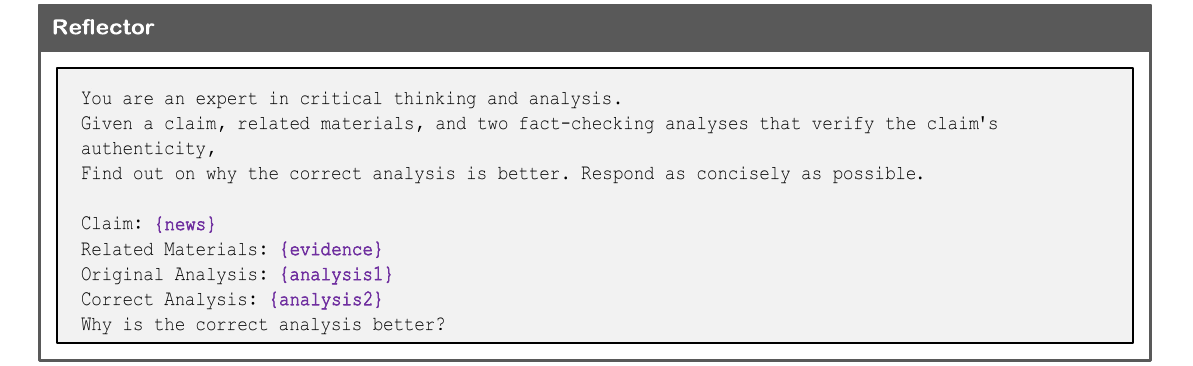}
  \includegraphics[width=1.0\textwidth]{images/appendix/Judger.pdf}
  \includegraphics[width=1.0\textwidth]{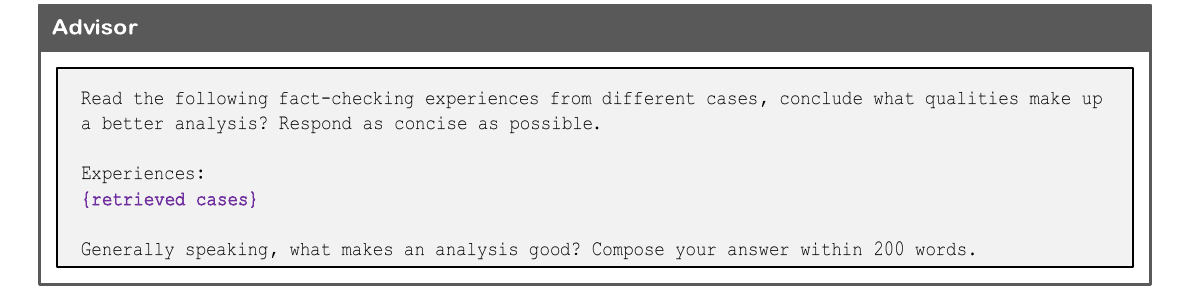}

  \caption{Prompt of \texttt{Reflector}, \texttt{Advisor}. and \texttt{Judger}}
  \label{fig:prompt_reflector}
\end{figure*}

\noindent\textbf{Dataset settings.}
We use the dataset setting of the previous methods. Specifically, 
for CHEF, we delete the label of ``NEI'' (Not Enough Information) from the dataset to make it a binary classification task following previous work.
All datasets are split into training, validation, and test sets in the ratio of 6:2:2. For RoE-FND we develop prompts on the training set and experiment on the testing set.
All datasets are downloaded from their official published sources:
CHEF~\url{https://github.com/THU-BPM/CHEF},
Snopes~\url{https://www.mpi-inf.mpg.de/dl-cred-analysis/},
PolitiFact~\url{https://www.mpi-inf.mpg.de/dl-cred-analysis/}.

\begin{figure}[!t]
  \centering
  \includegraphics[width=1.0\columnwidth]{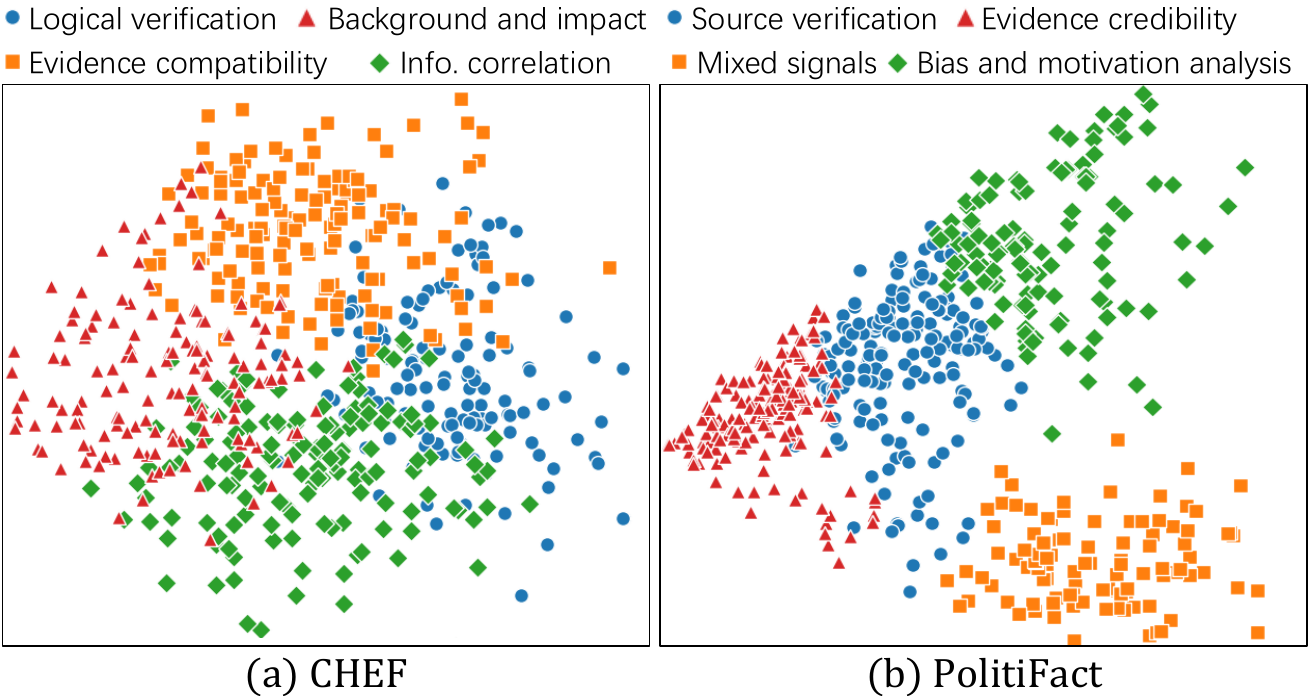}
  \caption{Visualizations of the \texttt{Analyzer}'s methodologies with keywords extraction and K-means clustering on Chinese and English datasets.}
  \label{fig:analyzer_generation}
\end{figure}


\noindent\textbf{Prompt design.}
In RoE-FND, we design the prompts in four parts: 1)~role assignment, 2)~output formation, 3)~task instruction, and 4)~input wrapper.
The prompts of ordinary and special \texttt{Analyst} are presented in Figure~\ref{fig:prompt_analyst}.
The prompts of \texttt{Reflector}, \texttt{Advisor}, and \texttt{Judger} are presented in Figure~\ref{fig:prompt_reflector}.

\section{Supplement Results}
\noindent\textbf{Generation of \texttt{Analyst}.}
The \texttt{Analyst} generates analysis about the news' authenticity. 
Although the conclusions may vary, authenticity analysis adheres to certain methodologies.
To uncover the underlying patterns of the methodologies, we engage an LLM to extract the keywords that anchor the methodologies.
Subsequently, we employ K-means clustering on the embeddings of the extracted keywords to identify several centroids. This process categorizes the methodologies into four distinct genres, as illustrated in Figure~\ref{fig:analyzer_generation}. The clustering outcomes demonstrate that the generated analyses are underpinned by robust logic across various dimensions. This comprehensive foundation ensures the reliability and interpretability of the proposed framework.

\noindent\textbf{Case Specific Criteria is Prefered than Common Criteria.}
The summarized criteria contain a list of keywords and their explanation, which are provided to the \texttt{Judger} as standard for it to select the better analysis.
We ask the \texttt{Judger} to score analyses by each criteria keyword. 
Then, it considers the score difference and selects the winning analysis. 
Therefore, the criteria keywords that distinguish the two analyses benefit the detection most.
We define the advantage of a criteria keyword as the difference between the score of the correct analysis and that of the incorrect one. 
This advantage is then normalized by dividing it by the keyword's occurrence count. Consequently, theoretically, criteria with greater advantages will have higher scores, capped at a maximum of ten.
In Fig~\ref{fig:judger_generation}, we visualize head advantageous criteria keywords and common keywords from Figure~\ref{fig:advice}. Notably, common criteria keywords offer limited advantages during the \texttt{Judger}'s selection process. The results underscore the importance of retrieving experiences from similar cases, as analyses may not be distinguishable based on common criteria alone.

\section{Licenses and Tterm for Use.}
All-MiniLM-L6-v2, Ollama uses MIT license. Table~\ref{tab:licenses} lists the utilized models' information. 
The above artifacts are utilized consistent with their intended use.


\end{document}